\documentclass[letterpaper,10pt,conference]{ieeeconf}
\IEEEoverridecommandlockouts
\usepackage{cite}
\usepackage{amsmath,amssymb,amsfonts}
\usepackage{algorithmic}
\usepackage{graphicx}
\usepackage{textcomp}
\usepackage{xcolor}
\usepackage{booktabs}
\usepackage{tabularx}
\usepackage{cuted}
\usepackage{multirow}
\usepackage{xspace}
\usepackage{colortbl}
\usepackage{svg}
\usepackage{balance}
\usepackage[most]{tcolorbox}

\usepackage{microtype} 

\usepackage[caption=false,font=footnotesize]{subfig}
\def\BibTeX{{\rm B\kern-.05em{\sc i\kern-.025em b}\kern-.08em
    T\kern-.1667em\lower.7ex\hbox{E}\kern-.125emX}}

\providecommand{\gap}[1]{\textbf{[GAP: #1]}}
\newcommand{\method}{VICTR\xspace}

\newcommand{\question}[2]{
    \vspace{-0.18cm}
    \begin{tcolorbox}[
        colback=white!90!gray,
        colframe=teal!60!black,
        arc=5pt, boxsep=3.5pt,
        left=10pt, right=10pt, top=1.5pt, bottom=1.5pt,
        boxrule=0.8pt,
        drop shadow=gray!50!white,
        enhanced jigsaw
    ]
    \vspace{-0.1cm}
        \noindent\textbf{\textit{Q#1:}}~#2
    \vspace{-0.1cm}
    \end{tcolorbox}
    \vspace{-0.1cm}
}

\newcommand{\NumEvalTasks}{16}

\newcommand{\NumRolloutsPerTask}{10}

\usepackage{hyperref}
\definecolor{inkblue}{HTML}{1F4E79}   
\hypersetup{colorlinks=true, urlcolor=inkblue, linkcolor=black, citecolor=black}

\begin{document}

\title{\LARGE\bfseries
VICTR: Value-Informed In-Context Learning for Vision-Language-Action Models
}

\author{Luca Macesanu$^{1,\dagger}$, Tianluo Zhang$^{1,\dagger}$, Aditya Patil$^{1}$, Archit Sharma$^{1}$, Kevin Angers$^{1}$,\\
Christopher M. Clark$^{1,\ddagger}$, Chen Feng$^{1,\ddagger}$, and Juexiao Zhang$^{1,\ddagger}$%
\thanks{\raggedright$^{\dagger}$Co-first authors. $^{\ddagger}$Co-advising. Correspondence: \texttt{\{juexiao.zhang, luca.m\}@nyu.edu}.}%
\thanks{$^{1}$New York University, New York, NY, USA.}%
}

\maketitle
\thispagestyle{empty}
\pagestyle{empty}




\begin{abstract}
Recent advances in robot foundation models have highlighted the potential of in-context learning (ICL) for adapting robot behavior from demonstrations without further fine-tuning. 
Previous work has used retrieval-based post-training to equip pretrained vision-language-action (VLA) models with the capability for ICL by teaching them to condition action predictions on retrieved visually-relevant demonstration chunks. However, the effectiveness of retrieval hinges on selecting context relevant to the current execution stage, which visual similarity alone may not capture.
We introduce VICTR, a framework enabling in-context learning in VLAs via progress-informed context retrieval built around an In-Context Value Function Estimator (IC-VFE). IC-VFE conditions on a reference demonstration video, a language instruction and the current observation to estimate task-relative \emph{progress}.
By conditioning progress estimation on a reference demonstration, IC-VFE can align the current state with a demonstrated execution even for tasks unseen during training.
VICTR uses the estimated progress to select demonstration chunks at corresponding stages of execution and supplies them to a $\pi_{0.5}$ policy post-trained for in-context action generation.
On unseen out-of-domain real-world manipulation tasks, VICTR raises the success rate from 6.2\% for a fine-tuned $\pi_{0.5}$ baseline to 23.8\% and more than doubles the task progress rate, from 28.9\% to 55.5\%. 
Project page with code, data and videos: \url{https://ai4ce.github.io/VICTR/}
\end{abstract}

\section{Introduction}
\label{sec:intro}

In-context learning (ICL) enables robot policies to adapt their behavior from demonstrations supplied as context, without repeatedly fine-tuning parameters for new tasks at deployment – mirroring large language models, which can infer new tasks from examples in their context without updating model parameters \cite{brown2020language}. The emergence of this capability in large pretrained models motivates an analogous question for robotics: whether generalist policies can similarly adapt their behavior from demonstrations supplied at inference time.
This form of adaptation can potentially address a persistent challenge for robotics foundation models. Despite substantial progress through training on diverse datasets~\cite{pi05} and reinforcement learning~\cite{pi06}, reliable out-of-the-box performance across diverse tasks and environments remains challenging~\cite{roboarena,fang2025intention}. 
Adapting a policy to a new domain often requires additional demonstration collection and fine-tuning, leading to a recurring cost as deployment requirements change. 
Recent industry systems have begun to demonstrate the promise of ICL, either as an emergent capability of large-scale pretraining~\cite{generalist2026gen15} or through pretraining objectives that explicitly use demonstration videos to specify tasks~\cite{skild2026s1}. 
Alongside these advances, RICL~\cite{ricl} demonstrates that pretrained VLAs can acquire ICL capabilities through retrieval-based post-training without retraining from scratch with extensive resources, learning to generate actions in-context conditioned on selected demonstration chunks.

\begin{figure}[t]
    \centering
    \includegraphics[width=\linewidth]{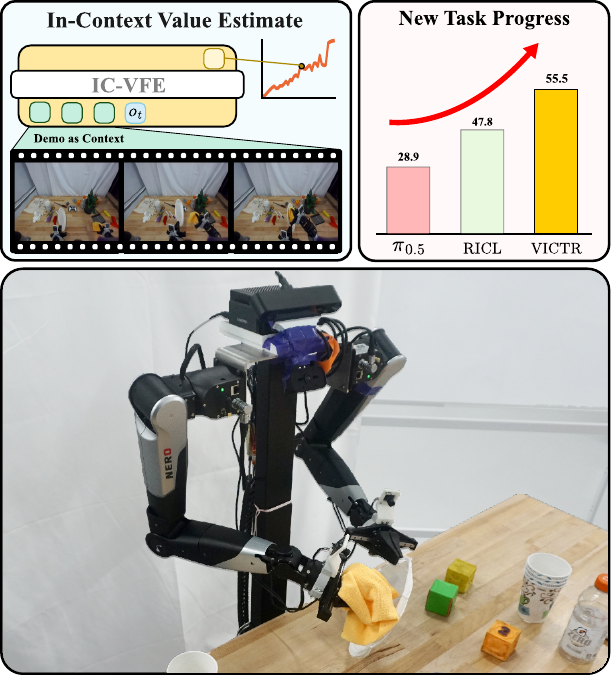}
    \caption{\textbf{Value-informed context retrieval for in-context robot learning.} Given a reference demonstration, IC-VFE estimates task-relative progress, which VICTR uses to retrieve progress-aligned context for action generation. On out-of-domain tasks, VICTR improves the task progress rate from 26.8\% with $\pi_{0.5}$ and 43.0\% with visual-similarity-based RICL to 58.2\%, without task-specific fine-tuning at deployment.}
    \label{fig:pull-fig}
\end{figure}


Retrieval-based ICL relies on selecting demonstration chunks that provide useful guidance for the robot's next action. Existing retrieval methods rely on visual similarity~\cite{ricl,mimicdroid}, but visually similar observations often occur at different execution stages, making appearance an ambiguous selection signal.
We argue that task-relative \emph{progress} provides an important retrieval signal beyond visual similarity, aligning the current observation with demonstration chunks at corresponding stages of execution.
A related notion appears in RECAP~\cite{pi06}, whose value function estimator (VFE) predicts the negative number of steps remaining until success as a distribution over discretised bins, and forms advantages from it for reinforcement learning. Normalised by the episode horizon, that value runs from $-1$ at the start of an episode to $0$ at success---the scale on which we define task-relative \emph{progress}, a unitless quantity in $[0,1]$ comparable across tasks of different lengths. We keep RECAP's formulation and this support, but supervise it with semantic progress annotations rather than elapsed time, and use the estimate to retrieve demonstration chunks at inference rather than to optimise the policy.
However, RECAP is conditioned on the current observation and language instruction without a reference demonstration, meaning that adaptation to new tasks depends on the VFE’s prior knowledge of the new language instruction.
We seek a progress estimator that can directly use demonstration context to interpret task-specific execution progress, without requiring task-specific fine-tuning.

To this end, we make progress estimation itself an in-context learning problem. We introduce an In-Context Value Function Estimator (IC-VFE) that jointly conditions on the current observation, a language instruction, and a reference demonstration video. The demonstration provides a task-specific reference \emph{in-context} for the task goal and how it unfolds, supplying execution information beyond the language instruction. IC-VFE uses this reference to assess the current state's progress relative to a demonstrated execution, enabling adaptation through context without task-specific fine-tuning. This estimator forms the core of \method, our \textbf{V}alue-\textbf{I}nformed \textbf{C}on\textbf{t}ext \textbf{R}etrieval framework for in-context VLA policies.

In \method, we instantiate IC-VFE with a Gemma-based~\cite{team2025gemma} vision-language architecture to predict task-relative progress, and post-train a $\pi_{0.5}$ policy to generate actions conditioned on demonstration chunks retrieved by progress alignment. Both models' weights remain fixed at deployment. In real-world evaluations, IC-VFE agrees with human progress
annotations more closely than existing online value estimators on both seen and unseen tasks, and this advantage carries through to control: on unseen out-of-domain tasks, \method{} raises the success rate of a fine-tuned
$\pi_{0.5}$ baseline from 6.2\% to 23.8\% and more than doubles
the task progress rate, from 28.9\% to 55.5\% (Figure \ref{fig:pull-fig}).

In summary, this work makes three primary contributions:
\begin{enumerate}
    \item An In-Context Value Function Estimator (IC-VFE) that uses a reference demonstration video to estimate task-relative progress without task-specific fine-tuning.
    
    \item \looseness=-1 VICTR, a $\pi_{0.5}$-based ICL framework that retrieves demonstration chunks by progress alignment to guide action generation.

    \item Real-world evaluations on 8 in-domain and 8 out-of-domain tasks, where progress-informed retrieval raises out-of-domain success from 2.7\% for a fine-tuned $\pi_{0.5}$ and 15.0\% for a visual-similarity retrieval baseline to 25.0\%.
\end{enumerate}

\section{Related Work}
\label{sec:related-work}

\textbf{In-context learning for embodied agents.} Prior work has explored injecting in-context adaptability into VLAs by retrieving and placing demonstration segments directly in a policy's context window~\cite{ricl,ravla}, learning single-demonstration ICL for non-VLA policies via next-token prediction~\cite{icrt}, and performing in-context imitation learning with explicit visual reasoning~\cite{iclr}.
Other work approaches ICL from unstructured human play data~\cite{mimicdroid}, studies how retrieved action consequences improve VLA generalization~\cite{reflectivevla}, or replaces sequence-conditioned retrieval with explicit correspondence matching between a single demonstration and the current scene~\cite{matchingpolicy}.
A parallel line of work scales context implicitly by compressing long interaction histories into online-updated fast weights via test-time training~\cite{robottt}; \method instead keeps policy weights fixed at test time and relies entirely on what is retrieved into the context window.
\method builds most directly on the retrieval-augmented formulation of RICL~\cite{ricl}, extending its visual-similarity-only retrieval with an in-context value estimate.

\textbf{Value and reward modeling for VLAs.} A growing body of work trains
general-purpose value/reward models for robot manipulation, including
stage-aware reward modeling~\cite{sarm} and other value estimators
built on VLM backbones~\cite{robometer,topreward}. RECAP~\cite{pi06} shows that a
distributional value function trained on episode-level success labels can
drive advantage-conditioned policy improvement at scale; our IC-VFE adopts a
similar distributional value-bin formulation
(Section~\ref{sec:ic-vfe-architecture}) but conditions the prediction on a
reference demonstration rather than the query observation alone. Closest in
spirit is VITA~\cite{vita}, which also targets out-of-domain value
calibration but does so by taking online gradient steps on a
self-supervised loss at test time; the IC-VFE instead conditions its progress
prediction on a reference demonstration, requiring no test-time parameter
updates. IC-VFE's targets are annotation-derived progress proxies rather than
discounted returns, and our evaluation measures prediction error rather
than probabilistic calibration.
Advantage-Weighted Flow Matching~\cite{awfm} similarly uses value annotations
to bias flow-matching-based policy training toward high-reward behavior.

\textbf{Retrieval in robot learning.} Beyond ICL specifically, retrieval
has been used to augment behavior-cloning policies with relevant offline data~\cite{mimicgen}, and dedicated retrieval-augmented VLA
architectures have been proposed for general manipulation~\cite{retrievalvla}.
LEACL~\cite{leacl} explores LLM-enhanced automatic curriculum construction for robot
learning. \method differs from this line of work in retrieving
based on a learned, task-progress-aware signal rather than similarity alone.


\section{Methodology}
\label{sec:methodology}

\begin{figure*}[t]
    \centering
    \includegraphics[width=0.98\textwidth]{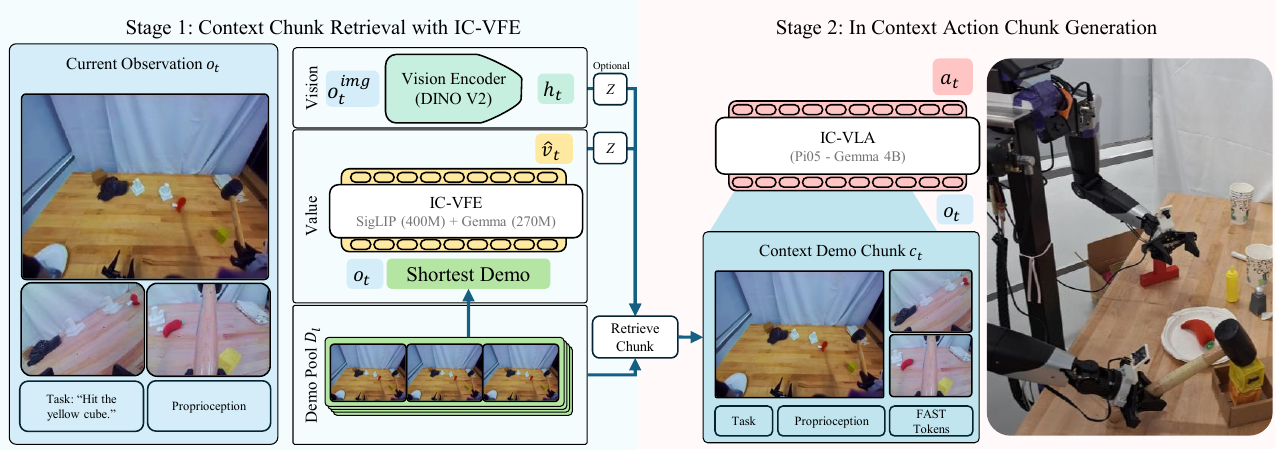}
    \caption{\textbf{\method method overview.}  The IC-VFE ($\phi$) conditions on the reference demonstration $\tau_{\text{ctx}}$,
    the shortest in the pool $\mathcal{D}_\ell$, to produce a progress estimate $\hat v_t$ for the current
    observation $o_t$. This is then optionally fused with visual similarity, driving
    retrieval of a context chunk $c_t$ from the demonstration pool. The IC-VLA
    ($\theta$) conditions on $c_t$ to predict the next action chunk,
    closing the loop without any parameter updates.}
    \label{fig:method}
\end{figure*}


\subsection{VICTR Overview}
\label{sec:method-overview}

\looseness=-1 We consider a multi-task vision-language-action setting. A task is specified by a
language instruction $\ell \in \mathcal{L}$, and an observation
$o_t = [X_t^1, \dots, X_t^n, q_t]$ consists of $n$ camera images and proprioceptive
state $q_t$. A VLA policy predicts action chunks $a_{t:t+H}$ conditioned on the instruction and observations.
\begin{equation}
    a_{t:t+H}  = \pi_\theta( o_t, \ell)
\end{equation}
In the following paragraphs we explain how $\pi_\theta$ is adapted for retrieval-based in-context learning.

\textbf{Demonstration Pool.}
At deployment, each task is provided with a pool of 10 \emph{context demonstrations} $\mathcal{D}_\ell$, each split into chunks aligned with the VLA action chunks. All demonstrations supply candidate chunks; the shortest demonstration additionally serves as the \emph{reference} $\tau_{\text{ctx}}$ for IC-VFE, detailed in Section~\ref{sec:ic-vfe-architecture}. 
Among successful demonstrations, shorter trajectories tend to contain fewer retries and corrective motions, making duration a simple proxy for demonstration quality and favoring a clean expert execution as the progress reference.
Let $c_1,\dots,c_M$ denote all the candidate chunks across all demonstrations in the pool, and let $o_i$ denote the observation at the start of chunk $c_i$. Conditioned on the task instruction and the shared reference demonstration, IC-VFE $f_\phi$ assigns each chunk a progress estimate:
\begin{equation}
\hat{v}_i
= f_\phi(o_i, \ell, \tau_{\text{ctx}}),
\label{eq:demo-value}
\end{equation}
where $\hat{v}_{i}\in[0,1]$ is the predicted progress.
For retrieval, the demonstration pool is indexed at the chunk level as a key-value store. Each chunk $c_i$ has a key consisting of its estimated progress $\hat{v}_i$ and DINOv2 visual embedding $h_i$. 
Its associated content consists of the three camera images and proprioceptive state at the start of the chunk, together with the FAST-tokenized~\cite{pertsch2025fast} action sequence. 
See Figure \ref{fig:method} for an overview of \method.

\subsection{Context Construction and Retrieval}
\label{sec:context-retrieval}

\textbf{Context Retrieval.}
At each inference step, we estimate the progress of the current observation $o_t$ using IC-VFE:
\begin{equation}
    \hat{v}_t = f_\phi(o_t,\ell,\tau_{\text{ctx}}),
    \label{eq:query-progress}
\end{equation}
where $\tau_{\text{ctx}}$ is the same reference demonstration used to compute progress for the demonstration pool. Our main method, \method, ranks the candidate chunks $c_1,\dots,c_M$ by their progress distance to the current observation:
\begin{equation}
    d_{\mathrm{val},i}
    = \left|\hat{v}_t-\hat{v}\right|.
    \label{eq:progress-distance}
\end{equation}
At each step we retrieve the single chunk with the smallest distance,
\begin{equation}
    i_t^\star = \arg\min_{i\in\{1,\dots,M\}} d_{\mathrm{val},i},
    \qquad c_t = c_{i_t^\star},
    \label{eq:retrieval-fn}
\end{equation}
and supply its observation and action sequence as context to the IC-VLA policy. We call the selected chunk a \emph{context chunk} when emphasising this role, and use the two terms interchangeably. Together with the current observation $o_t$ and instruction $\ell$, this context conditions the policy's next action-chunk prediction.

We also evaluate two alternative retrieval strategies using the same demonstration pool. \emph{Vision-only retrieval}, following the RICL recipe \cite{ricl}, ranks chunks by the distance between the current DINOv2 embedding $h_t=\Phi_{\mathrm{DINOv2}}(o_t)$ and each chunk's stored embedding $h_i$:
\begin{equation}
    d_{\mathrm{vis},i}
    = \left\|h_t-h_i\right\|_2.
    \label{eq:visual-distance}
\end{equation}
\emph{Vision+value retrieval} combines the two distances after normalizing each across the candidate chunks:
\begin{equation}
    g_i = z(d_{\mathrm{vis},i}) + z(d_{\mathrm{val},i}),
    \label{eq:fused_score}
\end{equation}
where
\begin{equation}
    z(d_i)=\frac{d_i-\mu_d}{\sigma_d+10^{-8}}.
    \label{eq:z_score}
\end{equation}
where $\mu_d$ and $\sigma_d$ are computed over the candidate pool separately for each distance. These variants replace $d_{\mathrm{val},i}$ in Equation~\ref{eq:retrieval-fn} with $d_{\mathrm{vis},i}$ or $g_i$, respectively.

\textbf{In-Context Conditioning of the VLA.} 
We use $\pi_{0.5}$ as our base VLA policy and prepend the tokenized representation of the retrieved demonstration chunk $c_t$ to the current query. The resulting input sequence has the form $[c_t,o_t,\ell]$, and is used to train the in-context policy $\pi_\theta^{\mathrm{ICL}}$ which predicts the next action chunk as
\begin{equation}
a_{t:t+H} = \pi_\theta^{\mathrm{ICL}}\big(c_t,o_t,\ell\big).
\label{eq:icl-vla}
\end{equation}
The demonstration block begins with visual tokens from three camera views---the head camera and the left and right wrist cameras---followed by proprioceptive state tokens. We encode and tokenize these observations using the same procedures as the original $\pi_{0.5}$ policy. These observation tokens are followed by FAST tokens representing the retrieved action chunk. To reduce context length, we retain only the lowest 20 of 30 discrete cosine transform frequency bins of the action chunk before FAST tokenization. The resulting action tokens are remapped to reserved vocabulary IDs.
The subsequent query block contains the current camera images, proprioceptive state, and language instruction. Block-causal attention allows query tokens to attend to the demonstration block while preventing demonstration tokens from attending to the query, enabling the policy to use the retrieved observation--action example when predicting $a_{t+H}$.

\subsection{In-Context Value Function Estimator}
\label{sec:ic-vfe-architecture}

IC-VFE estimates task-relative progress by conditioning on a reference demonstration. Given the current observation $o_t$, task instruction $\ell$, and reference demonstration $\tau_{\text{ctx}}$, the model outputs $\hat v_t=f_\phi(o_t,\ell,\tau_{\text{ctx}})\in[0,1]$. 
Here, we use \emph{value} to denote a task-relative progress signal rather than the conventional expected-return value $V^\pi$; IC-VFE is supervised using frame-level progress targets generated by Robo-Dopamine.
The reference demonstration is provided before execution; during rollout, IC-VFE has no access to future query observations or the terminal state.

\textbf{Context and Query Representation.}
We use the frozen vision encoder from Gemma 3 4B to process both current and reference observations. Each camera image is resized to $896\times896$ and encoded into 256 visual tokens after spatial pooling. To reduce the token and memory cost of encoding a full demonstration, we compress each reference view into 16 \emph{summary} tokens using two successive $2\times2$ patch-merging layers~\cite{bai2025qwen3}. For the current observation, we retain all 256 \emph{detail} tokens per view and additionally compute 16 summary tokens using the same patch-merging layers. This preserves fine visual information in the current observation while providing a shared representation for comparison with reference observations.
Both visual representations are mapped to the Gemma 3 270M decoder's 640-dimensional embedding space. A separate MLP projects each proprioceptive state into a single token. 
Let $\mathbf C_i$ denote the context block for the $i$-th sampled reference observation, containing summary tokens from all three camera views and the corresponding proprioceptive token. The current-observation block $\mathbf O_t$ contains both detail and summary visual tokens, together with its proprioceptive token. With instruction embeddings $\mathbf E_\ell$ and a learnable value-query token $\mathbf r$, the decoder input is
\begin{equation}
\mathbf H^{(0)}
=
[\mathbf C_1;\ldots;\mathbf C_K;
\mathbf O_t;\mathbf E_\ell;\mathbf r],
\label{eq:icl-vfe-seq}
\end{equation}
where the $K$ sampled reference observations are ordered chronologically.

\begin{figure}[t]
\centering
\includegraphics[
  width=\linewidth,
  trim={0pt 0pt 0pt 0pt},
  clip
]{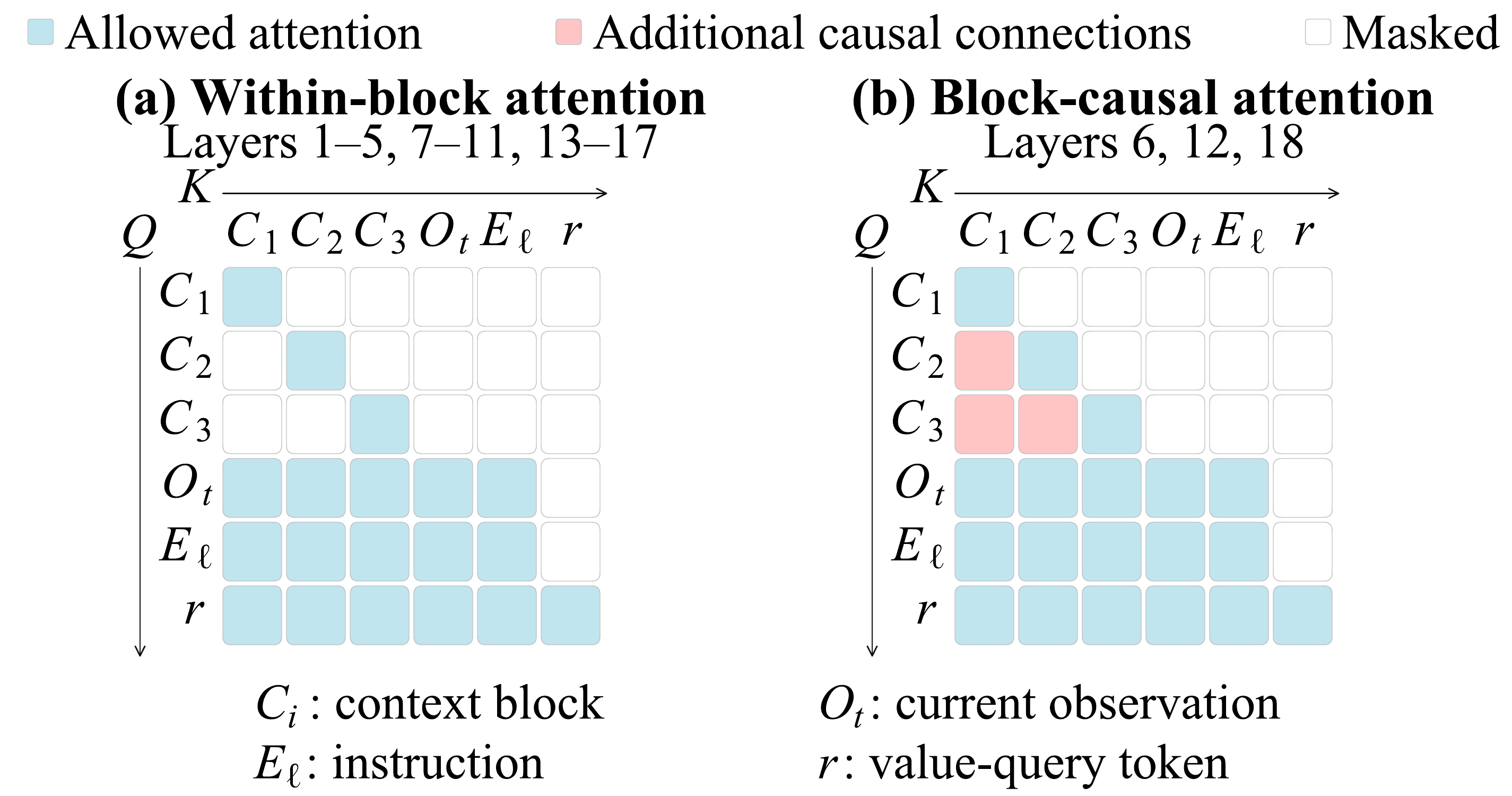}
\caption{\textbf{IC-VFE attention patterns}. Context blocks attend independently at most layers and communicate through block-causal attention at layers $h\in\{6,12,18\}$. The current-observation/instruction block and value-query token attend to the preceding context.}
\label{fig:block-causal-attention-mask}
\end{figure}

\textbf{Decoding with Hybrid Attention.}
An 18-layer Gemma 3 270M decoder processes this sequence using the two attention patterns illustrated in Figure~\ref{fig:block-causal-attention-mask}. In most layers, within-block attention is bidirectional within each timestep's context block and does not cross block boundaries. At layers $h\in\{6,12,18\}$, block-causal attention additionally allows each context block to attend to preceding blocks, enabling temporal aggregation across the demonstration. A learned block-index embedding is added to the tokens in each context block to encode its position within the reference sequence.
In every layer, the current-observation and instruction tokens attend bidirectionally within their shared block and to all valid context blocks. The value-query token attends to the entire valid sequence, while other tokens cannot attend to it. Context blocks cannot attend to the current query, and padding tokens are masked. An MLP head followed by softmax maps the final hidden state of the value-query token to a probability distribution over $B=201$ discrete bins.

\textbf{Progress prediction.}
We convert the predicted distribution over discrete bins into a continuous progress estimate for retrieval. The bins have uniformly spaced support values $\nu_b=-1+b/(B-1)$, where $b\in\{0,\ldots,B-1\}$. Let $p_{\phi,b}(o_t,\ell,\tau_{\text{ctx}})$ denote the probability assigned to bin $b$ after the softmax. The scalar is:
\begin{equation}
\begin{aligned}
\hat v_t
&= f_\phi(o_t,\ell,\tau_{\text{ctx}}) \\
&= 1+\sum_{b=0}^{B-1}
\nu_b \cdot p_{\phi,b}(o_t,\ell,\tau_{\text{ctx}}).
\end{aligned}
\label{eq:progress-predict}
\end{equation}
When normalized by the episode horizon, this value ranges from -1 at the start of an episode to 0 at success. We use the corresponding shifted scale, $[0,1]$, to define task-relative progress.

\textbf{Offline Progress Supervision.}
Elapsed time is an imperfect proxy for task progress, while manual progress annotations are costly to obtain at scale. We therefore use pseudo-labels generated by the pretrained General Reward Model of Robo-Dopamine~\cite{tan2025robo}, which is trained on large-scale data spanning real-world robotics, simulation, and human videos. The model processes the completed episodes, using each episode's final frame as a goal image to produce frame-aligned progress estimates. This configuration accesses future information unavailable during the corresponding rollout and therefore cannot be used directly for online observation scoring, but it makes it a privileged teacher to train IC-VFE.

We train IC-VFE to predict these teacher-generated targets from the current observation, instruction, and a reference demonstration available before execution. This provides progress supervision without requiring additional manual progress labels for our training data. Under \textbf{Q4} we assess the quality of these pseudo-labels by measuring their agreement with independent human progress annotations on our dataset.

\textbf{Training objective.}
\label{sec:icvfe-training}
Let $r_t$ denote the teacher-generated progress annotation expressed as a percentage. We normalize it to $v_t\in[0,1]$ and assign the nearest output bin:
\begin{equation}
\begin{aligned}
v_t &= \frac{\operatorname{clip}(r_t,0,100)}{100},\
b_t &= \operatorname{round}\big((B-1)v_t\big).
\end{aligned}
\label{eq：icvfe-supervision}
\end{equation}
For a minibatch of $N$ examples, we minimize categorical cross-entropy:
\begin{equation}
\mathcal L_{\mathrm{VFE}}(\phi)
=
-\frac{1}{N}\sum_{j=1}^{N}
\log p_{\phi,b_j}
\big(o_j,\ell_j,\tau_{\text{ctx}}^{(j)}\big).
\label{eq:icvfe-objective}
\end{equation}
For each query frame, we sample a reference demonstration from a different episode of the same task. We jointly optimize the visual adapters, proprioceptive encoder, learned embeddings, decoder, and value head using AdamW, while keeping the vision encoder frozen.

\subsection{IC-VLA Training Objective}
\label{sec:ic-vla-objective}
The IC-VLA is fine-tuned from the released $\pi_{0.5}$ official checkpoint following the official recipe, except that context is added. The action expert predicts
the continuous action chunk $a_{t:t+H}$ with the flow-matching loss.
During training, the context chunk $c_t$ is sampled using oracle progress values $v_t$ from
$\mathcal{D}_\ell^{\text{ctx}}$, which is held out from the training set and has zero overlap with the demonstration pool at evaluation time. At
deployment the query progress is estimated by the IC-VFE.
Errors in this estimate can affect retrieval, and their downstream impact is measured and discussed in the experiment section and in Figure \ref{fig:value-metric}.

\providecommand{\gap}[1]{\textbf{[GAP: #1]}}
\section{Experiments and Results}
\label{sec:experiments}

\textbf{Evaluation protocol and dataset.}
Our central claim concerns in-context adaptation to \emph{unseen} tasks,
which places two requirements on the evaluation protocol. First, the
relationship between each evaluation task and the meta-training tasks must
be explicitly controlled, so that generalization can be attributed to a
specific factor – novel objects, skill composition, or semantic
transfer – rather than conflated with memorization. Second, the evaluation trajectories and task configurations must be absent from our post-training data: public benchmarks
such as LIBERO~\cite{libero} and DROID~\cite{droid} offer broad task
coverage, but DROID is part of the pretraining corpus of
$\pi_{0.5}$-style models, and neither benchmark is organized around a
controlled seen/unseen factorization. We therefore collect our own
real-robot dataset, which additionally lets us control scene clutter and
object placement, and provides the freshly collected per-task
demonstration pools that our in-context protocol requires at evaluation
time.

The dataset consists of 28 tasks with 1{,}180 episodes totaling 7.5
hours: 7 hours for training, and 30 minutes of demonstrations that
form the demonstration pools for the 16 evaluation tasks. The
evaluation suite comprises 8 in-domain tasks (seen during training)
and 8 out-of-domain tasks constructed along the three factors above:
substituting novel objects into existing tasks, composing multiple learned skills, and introducing semantically similar but unseen tasks. For an OOD task, neither the IC-VLA nor IC-VFE is trained on that task; adaptation occurs solely through the task-specific demonstration pool provided at inference time.
All data collection and test rollouts use the real-robot setup described below. The supplementary video provides the task list; Appendix~\ref{app:libero100-icvla} reports LIBERO-100 results.

\textbf{Robot setup.}
We use a YOR bimanual robot~\cite{yor} in a stationary configuration, with a head-mounted ZED camera and a fisheye wrist camera on each of two
7-DoF AgileX Piper arms. The proprioceptive state $q_t$ is 18-dimensional,
comprising the 3D end-effector position and a 6D rotation
representation~\cite{rot6d} per arm, the latter chosen to avoid the
antipodal ambiguity of quaternions. The action space is 20-dimensional:
delta end-effector position, absolute 6D rotation, and a gripper command
for each arm. We use action chunks $a_{t:t+H}$ with $H=30$, so that each chunk spans one second at the 30\,Hz control rate used during data collection.

\textbf{Our analysis is organized around six questions:} the main result
(\textbf{Q1}), the VFE that produces the retrieval signal
(\textbf{Q2}--\textbf{Q4}), and two ablations on the retrieval setup itself
(\textbf{Q5},~\textbf{Q6}). 
 The three ablations that require them (\textbf{Q3}, \textbf{Q5} and~\textbf{Q6}) use a fixed subset of 4 in-domain and 4 out-of-domain tasks drawn from the 16 evaluation tasks. The OOD subset includes both novel-object tasks and tasks requiring unseen skill compositions, preserving multiple forms of distribution shift represented in the full evaluation suite. All other results use the full suite.

\label{sec:dataset}

\question{1}{Is value a good retrieval signal for in-context learning?}
\begin{figure}[t]
  \centering
  \includegraphics[width=0.95\linewidth]{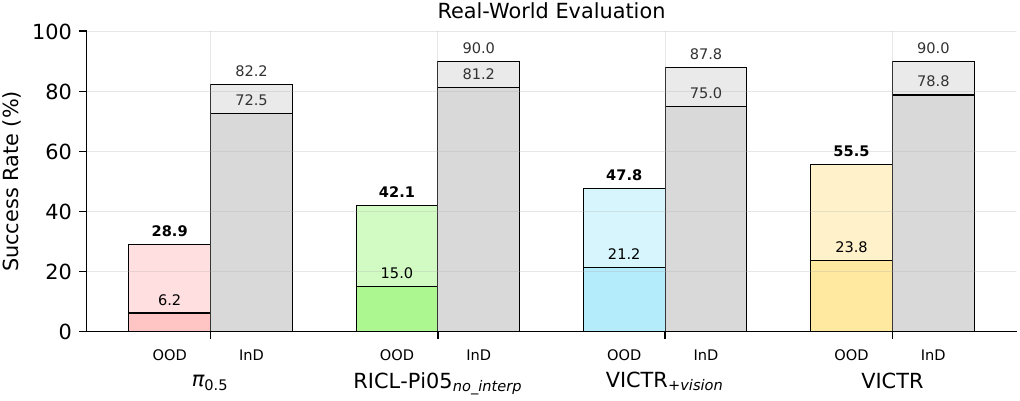}
  \vspace{-2mm}
  \caption{\textbf{Main result: success and task progress rates for $\pi_{0.5}$ and three retrieval-based variants on in-domain (InD) and out-of-domain (OOD) tasks.} Each bar's solid segment denotes binary success rate; the lighter segment adds partial credit for annotated milestones.}
  \label{fig:main-sr}
\end{figure}

\begin{figure*}[t]
    \centering
    \includegraphics[width=0.99\textwidth]{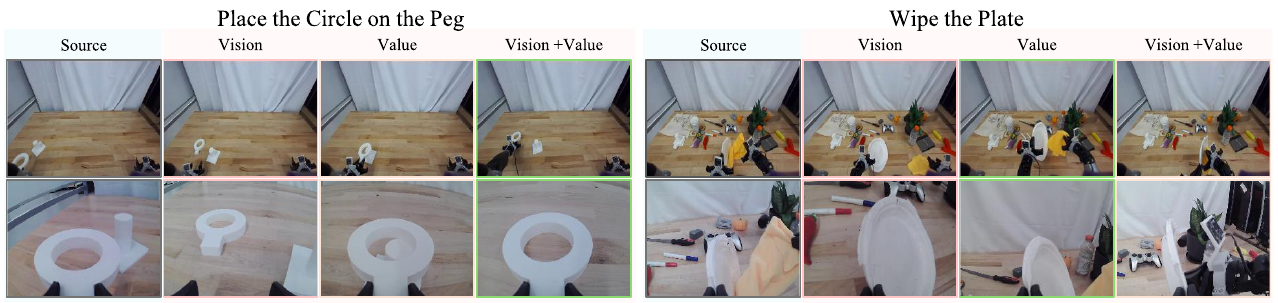}
    \vspace{-2mm}
    \caption{\textbf{Examples of retrieved context across retrieval signals.} The leftmost column on both sides ``source'' represents the query for the retrieval signal (and only the top camera is used during retrieval). For the circle-on-peg task, DINOv2 retrieves a suboptimal chunk, likely due to over-reliance on scene geometry. On the ``wipe the plate" task, IC-VFE retrieves a context chunk at a nearly identical stage of execution.}
    \label{fig:context-retrieval}
\end{figure*}

We first fine-tune a base $\pi_{0.5}$ model for 40k steps on our training data to adapt it to our robot and environment, similar to prior work that uses $\pi_{0.5}$-DROID~\cite{openpi} for the Franka setup. From this base checkpoint, all methods are allocated an additional 10k training steps.

As a non-ICL baseline, we continue fine-tuning $\pi_{0.5}$ without context. We additionally train two ICL baselines: RICL-$\pi_{0.5}$, which ports RICL's~\cite{ricl} implementation to $\pi_{0.5}$, and RICL-$\pi_{0.5,\mathrm{nointerp}}$, which removes its action interpolation from the original RICL. 
RICL was developed on $\pi_0$-FAST~\cite{pertsch2025fast}, not the more recent $\pi_{0.5}$.
Its interpolation layer blends the policy-predicted action with the action from the retrieved demonstration chunk, which can be unreliable when the retrieved demonstration state differs spatially from the current state, for example due to differences in object placement. Consistent with this, the interpolation layer did not transfer well to our setting: RICL-$\pi_{0.5}$ reaches only 0.5\% success, so we report the no-interpolation variant as the RICL baseline throughout. An ablation of this choice is conducted on LIBERO in Appendix~\ref{app:libero100-icvla}.
During training, the RICL variants retrieve context using DINOv2 visual similarity. VICTR uses offline progress annotations from Robo-Dopamine~\cite{tan2025robo} to supervise the IC-VFE and provide stable retrieval signal when training the IC-VLA. At deployment, VICTR retrieves with IC-VFE's estimated progress. Besides our main method VICTR, we additionally evaluate a variant VICTR$_{\mathrm{+vision}}$ that adds visual similarity to the retrieval signal.

All methods are evaluated over \NumRolloutsPerTask{} rollouts per task on all \NumEvalTasks{} tasks. 
Figure~\ref{fig:main-sr} reports task success rate and \emph{task progress rate}.
Task progress is defined using a task-specific sequence of $n$ predefined milestones, with completion of milestone $k$ corresponding to progress $k/n$; the full annotation protocol is described under Q4. We report both throughout: binary success is the primary metric, while progress rate additionally credits partial completion, especially on out-of-domain tasks where success rates are generally lower.

Retrieval-based methods provide their largest gains on out-of-domain tasks. The fine-tuned $\pi_{0.5}$ baseline achieves 6.2\% binary success and 28.9\% progress on OOD tasks. RICL-$\pi_{0.5,\mathrm{nointerp}}$, which retrieves context using visual similarity, improves performance to 15.0\% success and 42.1\% progress, demonstrating the benefit of retrieval-based in-context conditioning over the non-ICL baseline. 
VICTR retrieves with task progress instead of visual similarity and yields a further improvement, to 23.8\% success and 55.5\% progress; the wider margin in progress rate indicates that stage-aligned context carries the policy further through a task even when it does not finish it.
Finally, VICTR$_{\mathrm{+vision}}$, which fuses the two signals, reaches 21.2\% success and 47.8\% progress and therefore does not improve upon VICTR.
We hypothesize that equal weighting of the two normalized distances allows noisy visual similarity to override otherwise informative progress alignment. 
A potentially beneficial combination of both is left to future exploration.

On in-domain tasks, the distinction is less pronounced. All retrieval-based methods achieve high progress rates of 87.5\%-90\%, compared with 82.2\% for $\pi_{0.5}$. And all of them achieve high task success rates. 
The gains that do arise for VICTR over the baselines tend to occur on underrepresented semi-dexterous tasks, such as passing an object from one arm to the other or precisely placing a circle on a peg.

These results suggest that the choice of retrieval signal becomes particularly important when the policy must generalize beyond its training distribution, which is exactly where in-context-learning is meant to help. Visual similarity presents two challenges for retrieval. First, visually similar observations may correspond to different stages of execution and therefore different actions, as observed in Figure \ref{fig:context-retrieval}. Second, visual similarity can be sensitive to task-irrelevant changes in the environment. This is particularly relevant in our evaluation, where we deliberately alter distractor objects and local background appearance between data collection and evaluation, thereby changing the visual features used for retrieval. Progress provides a more task-centric notion of relevance, selecting examples according to \emph{where the robot is in the demonstrated task} rather than appearance alone.

\question{2}{Does context improve online value estimation?}

\setlength{\tabcolsep}{1pt}
\begin{table}[t]
  \centering
  \caption{\textbf{VFE quality: agreement with human-annotated progress curves}, on the \NumEvalTasks{} evaluation tasks.}
  \label{tab:vfe-results}
  \footnotesize
  \begin{tabular}{lcccc}
    \toprule
    & \multicolumn{2}{c}{In-domain} & \multicolumn{2}{c}{Out-of-domain} \\
    \cmidrule(lr){2-3} \cmidrule(lr){4-5}
    Method & Kendall $\tau_b$ $\uparrow$ & Pearson $r$ $\uparrow$ & Kendall $\tau_b$ $\uparrow$ & Pearson $r$ $\uparrow$ \\
    \midrule
    \multicolumn{5}{l}{\textit{Online value models}} \\
    \textbf{IC-VFE (ours)}                    & \textbf{0.51} & \textbf{0.62} & \textbf{0.52} & \textbf{0.65} \\
    \quad no context         & 0.26          & 0.36          & 0.38          & 0.49 \\
    \quad no block embedding  & 0.46          & 0.55          & 0.37          & 0.45 \\
    RECAP \cite{pi06}        & 0.48          & 0.60          & 0.43          & 0.57 \\
    Robometer \cite{robometer}     & 0.26          & 0.34          & 0.27          & 0.38 \\
    Robometer FT    & 0.32          & 0.44          & 0.33          & 0.46 \\
    \bottomrule
  \end{tabular}
\end{table}
\setlength{\tabcolsep}{6pt}

Adding a full demonstration into the context is our design choice in IC-VFE. Does it help?
Table~\ref{tab:vfe-results} scores each model against manually annotated
progress curves by Pearson correlation $r$ and Kendall's $\tau_b$; the
annotation protocol is described under \textbf{Q4}. We compare our
IC-VFE (Section~\ref{sec:ic-vfe-architecture}) against RECAP~\cite{pi06} and
Robometer~\cite{robometer}. Both have previously been used to provide reward signals for
real-world RL. We evaluate
Robometer both zero-shot and after fine-tuning on the training set's
sparse reward annotations. IC-VFE achieves
the best Kendall $\tau_b$ and Pearson $r$ among all online candidates
on both in-domain and out-of-domain tasks. 

The ablations also confirm that IC-VFE uses the reference context: removing context substantially reduces both correlation metrics in Table I. The block-index embedding introduced in Section \ref{sec:ic-vfe-architecture} provides an additional improvement.

\begin{figure}[t]
  \centering
  \includegraphics[width=0.95\linewidth]{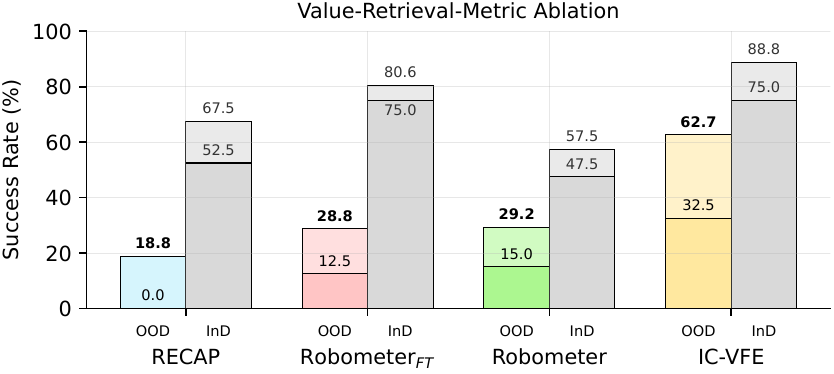}
  \vspace{-2mm}
  \caption{\textbf{VFE at deployment: task progress rate by value estimation method.} All methods are evaluated using the same IC-VLA backbone using Robo-Dopamine annotations at training time. IC-VFE yields substantially higher OOD progress than the alternative online estimators, indicating that retrieval quality is sensitive to the value estimator used at deployment.}
  \label{fig:value-metric}
\end{figure}

\question{3}{How does VFE quality affect task performance?}

Table~\ref{tab:vfe-results} identifies IC-VFE as the best performing online
VFE by offline agreement with human annotations. To test whether improved value estimation translates to a better policy, we evaluate \method on the ablation task set with IC-VFE,
Robometer, Robometer (FT) and $\pi_{0.6}$ RECAP each supplying the retrieval
signal, with the IC-VLA backbone fixed (Figure~\ref{fig:value-metric}).

Deploying \method with IC-VFE reaches 62.7\% out-of-domain progress rate, against 29.2\% for the best alternative, consistent with the offline evaluation in Table~\ref{tab:vfe-results}. Interestingly, fine-tuning Robometer on our data degraded its out-of-domain performance, despite the expected benefit of adapting it to our embodiment and environment.

Together, Q2 and Q3 show that improved progress estimation translates to improved retrieval-conditioned control. Figure~\ref{fig:context-retrieval} shows qualitative examples of retrieved context under each retrieval signal.

\question{4}{Which offline value model should supervise VFE training?}

\setlength{\tabcolsep}{1pt}
\begin{table}[t]
  \centering
  \caption{\textbf{Reward model quality: agreement with human-annotated progress curves} for offline value models used to produce VFE training signal, on the \NumEvalTasks{} evaluation tasks. \\FT: fine-tuned in-domain.}
  \label{tab:reward-results}
  \footnotesize
  \begin{tabular}{lcccc}
    \toprule
    & \multicolumn{2}{c}{In-domain} & \multicolumn{2}{c}{Out-of-domain} \\
    \cmidrule(lr){2-3} \cmidrule(lr){4-5}
    Method{\renewcommand*{\thefootnote}{\fnsymbol{footnote}}\footnotemark[1]}
      & Kendall $\tau_b$ $\uparrow$ & Pearson $r$ $\uparrow$ & Kendall $\tau_b$ $\uparrow$ & Pearson $r$ $\uparrow$ \\
    \midrule
    Robo-Dopamine \cite{tan2025robo}          & \textbf{0.79} & \textbf{0.88} & 0.79 & \textbf{0.90} \\
    Robo-Dopamine FT                          & 0.64          & 0.72          & \textbf{0.81}          & 0.88 \\
    Robometer \cite{robometer}                & 0.56          & 0.70          & 0.58          & 0.73 \\
    Robometer FT                              & 0.65          & 0.78          & 0.68          & 0.81 \\
    TOPReward\footnotemark[2] \cite{topreward} & 0.80          & 0.74          & 0.81          & 0.79 \\
    GVL \cite{gvl}                            & 0.34          & 0.41          & 0.29          & 0.37 \\
    \bottomrule
  \end{tabular}
\end{table}
\setlength{\tabcolsep}{6pt}

\footnotetext[2]{TOPReward returns degenerate values which increase monotonically with elapsed time regardless of the task progress.}

{\renewcommand*{\thefootnote}{\fnsymbol{footnote}}
Among the offline\footnote{Offline value models require information from
beyond the current timestep -- e.g.\ Robo-Dopamine takes the episode's final
frame as a goal image -- and cannot run during a real-world rollout.}
value models (Table~\ref{tab:reward-results}), Robo-Dopamine
(zero-shot) achieves the strongest agreement with the manual annotations, and thus we adopt it as the training signal for IC-VFE, Robometer and RECAP.
We score every candidate
value model against manually annotated progress curves. For each task, a task-specific list of $n$ subtask milestones is
defined in advance; marking
the $k$-th milestone assigns it a progress value of $k/n$ on a 0--1 scale,
so that the value awarded for a given point in a task follows a fixed per-task rubric rather than ad hoc annotator judgment. Milestones correspond to discrete, semantically meaningful stages of task completion and are defined from the task specification before evaluating any policy rollouts. Failures are
annotated by marking progress back down to whichever milestone is still
satisfied. Each value is eased in over a smoothstep function, with an
empirically chosen 2.5s lead-in window, to avoid discontinuities in the
value function.
Candidate models are scored against the reference progress curve with two
complementary statistics: Pearson correlation
$r$, and Kendall's $\tau_b$~\cite{topreward}.}

\begin{figure}[t]
  \centering
  \includegraphics[width=0.95\linewidth]{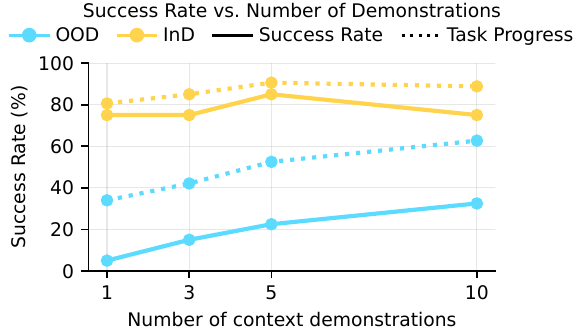}
  \vspace{-2mm}
  \caption{\textbf{Retrieval ablation: success rate vs.\ number of context demonstrations.} We evaluate \method with a variable number of demonstrations in its demonstration pool. Performance generally improves as the demonstration pool grows, saturating around five demonstrations.}
  \label{fig:demo-count}
\end{figure}
\question{5}{How many context demonstrations are needed?}

Using \method, we vary the number
of demonstrations in the pool over $\{1, 3, 5, 10\}$. For a pool of size $k$, we select the $k$ shortest demonstrations (Figure~\ref{fig:demo-count}).

Performance generally improves with pool size, consistent with larger pools providing more candidate chunks from which to retrieve relevant context.

\begin{figure}[t]
  \centering
  \includegraphics[width=0.95\linewidth]{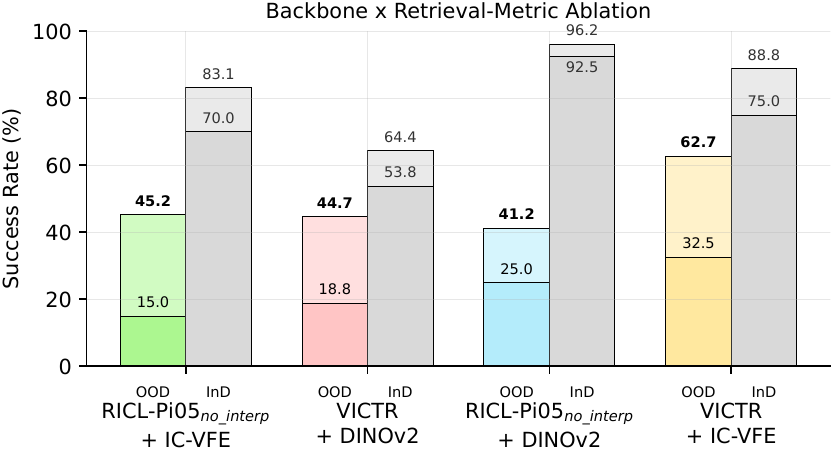}
  \vspace{-2mm}
  \caption{\textbf{Retrieval ablation: retrieval signal at training vs.\ deployment.} Labels at the bottom represents: \textit{the method at training} ``+" \textit{the retrieval signal at deployment}. While policies evaluated with the same retrieval signal they were trained with perform better across the board, it is not enough to overcome a poor retrieval signal at deployment time.}
  \label{fig:retrieval-mismatch}
\end{figure}
\question{6}{Must training and deployment use the same retrieval signal?}
We take \method and RICL-$\pi_{0.5,\mathrm{nointerp}}$ and test each with both retrieval signals, IC-VFE value and DINOv2 visual similarity (Figure~\ref{fig:retrieval-mismatch}). Results indicate that each policy performs best with the retrieval signal it was trained with, but that both still exhibit meaningful performance when swapped. 
We hypothesize that RICL-$\pi_{0.5,\mathrm{nointerp}}$ degrades under IC-VFE retrieval because it was trained on the distribution of chunks produced by DINOv2 retrieval, which may shape how the policy interprets its context.

\section{Limitations and Future Work}
\label{sec:limitations}

\method{} has three main limitations. First, although absolute performance on
unseen tasks improves substantially over the baselines, it remains low
(25.0\% success), leaving considerable room before in-context adaptation is
practically reliable. Second, all data collection and evaluation take place
in a single physical environment: while we vary objects and their placement, these local appearance changes do not constitute evaluation in a genuinely novel environment. Third, the context length of the IC-VFE is a
bottleneck when encoding the reference demonstration: the aggressive spatial
pooling required to fit a full demonstration video into context sacrifices
spatial resolution, which may limit progress estimation on tasks where
fine-grained visual detail matters.

Future work could relax the IC-VFE context budget through efficient attention or test-time-training techniques~\cite{robottt}, extend reference conditioning to human demonstration videos~\cite{wang2025vlm}, and evaluate progress-informed retrieval across more diverse environments.

A related question is whether \method{} remains relevant as the VLA formulation itself evolves. 
Recent systems depart from it in different directions: world action models such as DreamZero~\cite{wam} 
fine-tune a video diffusion backbone rather than a VLM, 
and general-purpose multimodal models such as Astra~\cite{gpt6astra} are beginning to be applied to
embodied control. 
We are also very intrigued by Astra's potential and conducted a pilot study of it 
under our evaluation setting. 
Our results and discussion are presented in Appendix~\ref{app:astra}.
While these advances are exciting and encouraging, 
we believe robotics remains a physically grounded domain,
in which adapting in-context to the given environment will 
remain an important capability regardless of how the underlying models and architectures evolve.
We build on $\pi_{0.5}$ because it remains the most widely used open base policy, but \method{}'s contribution – using estimated task-relative progress to select \emph{which} demonstration belongs in context – is not tied to it. 
Any policy that adapts in-context from demonstrations placed in its input faces the same selection problem, arguably more acutely once context is shared between language and action tokens. Porting progress-informed retrieval to these backbones is a
natural next step.

\section{Conclusion}
\label{sec:conclusion}

This work introduces the In-Context Value Function Estimator (IC-VFE) and \method, a framework that retrieves demonstration context for VLA policies by estimated task-relative progress rather than visual similarity. Conditioning
progress estimation on a reference demonstration allows a single fixed
estimator to interpret execution progress on tasks unseen during training, without task-specific fine-tuning. On real-robot out-of-domain tasks,
\method{} substantially outperforms both a fine-tuned non-ICL baseline and visual-similarity retrieval, and IC-VFE tracks human-annotated progress more closely than existing online value estimators – an advantage that carries through to deployment, where policies driven by IC-VFE retrieval achieve the highest out-of-domain progress among all estimators tested.
These results establish task-relative progress as an effective retrieval signal for in-context VLA adaptation, with the remaining gaps discussed in
Section~\ref{sec:limitations}.

\section*{Acknowledgment}
We thank Qinxi Yu, Kamil Imam and Greta Perez-Haiek for helpful discussions. The work was supported in part through NSF grants 2238968, and by NYU IT High Performance
Computing resources, services, and staff expertise. This research was also supported by the NVIDIA RTX 6000 Ada GPUs awarded to the AI4CE lab via the NVIDIA Academic Grant Program.

\balance
\bibliographystyle{IEEEtran}
\bibliography{references}

\clearpage
\nobalance

\appendix
\balance

\subsection{IC-VLA Experiments on LIBERO-100}
\label{app:libero100-icvla}
Our main experiments and results focus on real-world robot data and rollout, 
and we also release our data for public use.
But here we also evaluate on 30 unseen LIBERO-100 tasks~\cite{libero},
as a supplementary on a popular and accessible benchmark.
We compare non-ICL baselines, RICL with and without action interpolation,
and \method (Table~\ref{tab:libero100-icvla}). The $\pi_0$-FAST runs
use a 70/30 seen/unseen split with seed 123. All $\pi_{0.5}$ models
are trained for 10k steps with batch size 32, with retrieval top-$k$ variants
using $k\in\{1,4\}$ context chunks.

\begin{table}[ht]
    \centering
    \caption{\textbf{IC-VLA success rates on 30 unseen LIBERO-100 tasks.}
    Bold denotes the best result within each backbone.}
    \label{tab:libero100-icvla}
    \footnotesize
    \setlength{\tabcolsep}{5pt}
    \begin{tabular}{lccr}
        \toprule
        Method & Top-$k$ & Action interp. & Success (\%) \\
        \midrule
        \multicolumn{4}{l}{$\pi_0$-FAST backbone} \\
        \midrule
        Non-ICL baseline & -- & -- & 21.467 \\
        RICL & -- & Yes & \textbf{32.600} \\
        RICL & -- & No & 32.067 \\
        \midrule
        \multicolumn{4}{l}{$\pi_{0.5}$ backbone} \\
        \midrule
        Non-ICL baseline & -- & -- & 37.800 \\
        RICL & 1 & Yes & 26.200 \\
        RICL & 4 & Yes & 26.600 \\
        RICL & 1 & No & 41.200 \\
        RICL & 4 & No & 37.270 \\
        \method & 1 & No & \textbf{45.870} \\
        \method & 4 & No & 39.730 \\
        \bottomrule
    \end{tabular}
\end{table}

\textbf{Effect of action interpolation.}
Both RICL variants improve over the $\pi_0$-FAST baseline, 
and action interpolation produces a slightly better success rate.
This observation is consistent with their original paper.
However, with $\pi_{0.5}$ as backbone, removing
interpolation improves RICL from 26.200\% to 41.200\% at $k=1$ and
from 26.600\% to 37.270\% at $k=4$, consistent with our observations on the real-robot
results in Section~\ref{sec:experiments}. Therefore, 
action interpolation is not used in our main experiments.

\textbf{Retrieval strategy and context count.}
\method has the best performance among all methods evaluated.
\method achieves the best success rate of 45.870\% at $k=1$,
exceeding the $\pi_{0.5}$ baseline by 8.07 percentage points and RICL
without interpolation by 4.67 points. It also outperforms RICL without
interpolation at $k=4$. Both methods perform better with one retrieved
chunk than with four.

\subsection{Pilot Study: GPT-6 Astra as an In-Context Policy}
\label{app:astra}

As this work was being finalized, 
GPT-6 Astra~\cite{gpt6astra} was released and has shown strong generalization on robotic tasks. 
Therefore, we ran a pilot study replacing the \method{} policy with Astra under our evaluation setting: 
Astra receives the language instruction, live images from the head and both wrist cameras, 
and the measured arm states, and controls the arms directly through delta end-effector poses. 
Accessed through the Codex CLI, Astra first writes its own executor, a long-running process that 
connects to the robot over the same command channel as the \method{} inference pipeline, 
streams end-effector targets at 30\,Hz and enforces safety stops; this harness is then frozen 
for all attempts. At run time, Astra reasons about the scene and the task phase, 
and submits a short (1--3\,s) action chunk, which the executor runs, 
then Astra collects new observations and plans for the next action chunk.
Directly replaying demonstrations is prohibited. We compare two conditions: \emph{ICL},
in which Astra is given the same 10 demonstrations of the task that \method{} retrieves from 
and must cite the demonstration segments that informed each chunk, and \emph{zero-shot}, 
with no demonstrations. We evaluate the simplest and the hardest task in our suite, 
\textit{putting the square on the peg} and \textit{sorting the items into their containers};
since Astra is not trained on our data, both are out-of-domain for it. 
We run 5 attempts per task and condition, each in a fresh session. 
Scoring follows our protocol; for sorting without demonstrations, an item counts as sorted if it ends up in any container.

On putting square on peg, Astra reaches 20\% success and 50\% progress with ICL, 
versus 20\% and 40\% zero-shot; on sorting, it reaches 0\% success and 4.4\% progress in both conditions. 
Qualitatively, Astra shows strong visual and spatial generalization: 
it always correctly identifies the task-relevant objects and their roles,
understands arm motion and spatial directions on an embodiment it was never trained on, 
and, even without demonstrations, makes purposeful progress toward the goal. 
It also recovers from disturbances: 
when contact with the square pushed the peg away from its original position, 
Astra noticed the displacement and immediately reached further to resume the insertion. 
Its weaknesses are physical. Inference through the API is slow, 
and fine-grained control remains limited, e.g., 
choosing grasp and placement poses suited to an object's shape, 
or adjusting heights and distances to avoid collisions.

Notably, we observe that demonstrations help Astra less than they help $\pi_{0.5}$ in \method{}. 
We hypothesize that in-context demonstrations mainly help $\pi_{0.5}$ generalize to unseen tasks, 
whereas Astra already plans what an unseen task requires from the current observation and 
instruction alone. 
Our demonstrations are also all successful, 
with object placements that differ from the test scene. 
More useful context for a model like Astra may be its own attempts
in the same environment with success or failure feedback. 
So we believe recursive self-improvement from in-context experience as a promising direction,
alongside improving physical perception and distilling frontier models into 
smaller LLMs for local, low-latency deployment. 
Overall, Astra is already a capable high-level reasoner and planner for manipulation, 
while physical execution remains the bottleneck where robot-specific policies currently hold the advantage.

\subsection{Evaluation Task Suite}
\label{app:eval-tasks}

We evaluate on 16 real-world bimanual tasks, summarized in
Table~\ref{tab:eval-tasks}. The 8 \emph{in-domain} tasks are seen during
training. The 8 \emph{out-of-domain} tasks are unseen during training, and
each departs from the training distribution in one controlled way:
\emph{novel object} tasks reuse a trained skill on an object or goal never
paired with it in training; \emph{skill composition} tasks chain two trained
skills into a single instruction; the \emph{similar unseen task} requires an
extension of a trained skill; and the \emph{long-horizon} task requires
completing a multi-step sequence.

\definecolor{tagnovel}{HTML}{2F5FC4}
\definecolor{tagcomp}{HTML}{7E4BB8}
\definecolor{tagsimilar}{HTML}{C8790F}
\definecolor{taglong}{HTML}{1F5E66}
\newtcbox{\tasktag}[1][tagnovel]{on line, colback=#1, colframe=#1,
    coltext=white, boxrule=0pt, arc=3pt, boxsep=0pt,
    left=3pt, right=3pt, top=1pt, bottom=1pt,
    fontupper=\scriptsize\bfseries}

\begin{table}[h]
    \centering
    \caption{Evaluation tasks. In-domain tasks are seen during training;
    out-of-domain tasks are unseen and tagged by how they differ from the
    training distribution.}
    \label{tab:eval-tasks}
    \footnotesize
    \renewcommand{\arraystretch}{1.2}
    \begin{tabular}{@{}ll@{}}
        \toprule
        \multicolumn{2}{@{}l}{\textbf{In domain} \textit{(seen in training)}} \\
        \midrule
        \multicolumn{2}{@{}l}{Clean the plate} \\
        \multicolumn{2}{@{}l}{Hit the yellow cube with the mallet (right arm)} \\
        \multicolumn{2}{@{}l}{Pass the red chilli from left to right arm} \\
        \multicolumn{2}{@{}l}{Put the eggplant on the plate (left arm)} \\
        \multicolumn{2}{@{}l}{Place the red chilli in the cup} \\
        \multicolumn{2}{@{}l}{Put the circle on the peg} \\
        \multicolumn{2}{@{}l}{Put the eggplant into the box (left arm)} \\
        \multicolumn{2}{@{}l}{Uncap the blue marker} \\
        \midrule
        \multicolumn{2}{@{}l}{\textbf{Out of domain} \textit{(unseen in training)}} \\
        \midrule
        \tasktag[tagnovel]{novel object}           & Put the square on the peg \\
        \tasktag[tagnovel]{novel object}           & Uncap the black marker \\
        \tasktag[tagnovel]{novel object}           & Put the mustard into the box \\
        \tasktag[tagnovel]{novel object}           & Hit the eggplant with the mallet \\
        \tasktag[tagcomp]{skill composition}       & Pass the mustard to the plate \\
        \tasktag[tagcomp]{skill composition}       & Stack a cube in the box, then hit it \\
        \tasktag[tagsimilar]{similar unseen task}  & Stack three cubes into a tower \\
        \tasktag[taglong]{long horizon}            & Sort the items into their containers \\
        \bottomrule
    \end{tabular}
\end{table}

\subsection{Author Contributions}
\label{app:contributions}
All authors made essential contributions to the work. Below is a detailed description of  author contributions.

\textbf{Luca Macesanu} led the project; proposed and implemented the VICTR architecture; developed and trained the IC-VLA and baselines on real-world data;  designed the data distribution and experiments; and created most of the figures.

\textbf{Tianluo Zhang} led the design and development of IC-VFE, including its dual-resolution inputs and its block embedding; designed and conducted its ablations; trained and evaluated VICTR and RICL on LIBERO.

\textbf{Aditya Patil} collected and annotated a large portion of the data, ran the real robot rollouts for all methods and GPT6 Astra, and developed a majority part of the deployment infrastructure.

\textbf{Archit Sharma} set up and managed the evaluations of the value function estimators and managed the generation of the value annotations used to train the final IC-VFE. Designed and implemented the project page.

\textbf{Kevin Angers} contributed significantly to the training and deployment infrastructure. Diagnosed and fixed critical failures that blocked policy training and real-world deployment. Contributed to writing the paper.

\textbf{Christopher Clark} advised the project, providing feedback on the approach and raising questions that helped guide the work.

\textbf{Chen Feng} advised the project, his guidance in framing the work and identifying contributions was essential.

\textbf{Juexiao Zhang} served as the project's main advisor, meeting with the team weekly to guide its direction and methodology development, led the writing of the manuscript.

\end{document}